\documentclass{uncecomp_preprint}
\usepackage[utf8]{inputenc}
\usepackage[T1]{fontenc}

\usepackage{amsmath, amssymb, bm} 
\usepackage{array}
\usepackage{graphicx}
\usepackage{textcomp} 
\usepackage{caption}
\usepackage{subcaption}

\usepackage[backend=biber, style=numeric, sorting=none, isbn=false, url=false]{biblatex}
\AtEveryBibitem{\clearfield{month}}
\AtEveryBibitem{\clearfield{day}}

\AtEveryBibitem{%
	\ifentrytype{online}{
		\clearfield{url}%
		\clearfield{urlyear}%
	}{}
}

\title{Fourier Neural Operator surrogate model to predict 3D seismic waves propagation}

\author{Fanny Lehmann$^{1,2}$, Filippo Gatti$^2$, Michaël Bertin$^1$, and Didier Clouteau$^2$}

\heading{F. Lehmann, F. Gatti, M. Bertin, and D. Clouteau}

\address{$^1$CEA/DAM/DIF\\
  F-91297, Arpajon, France\\
  e-mail: michael.bertin@cea.fr \and
  $^2$
  Université Paris-Saclay, ENS Paris-Saclay, CentraleSupélec, CNRS, LMPS - Laboratoire de Mécanique Paris-Saclay \\
  91190 Gif-sur-Yvette, France\\
  e-mail: \{fanny.lehmann,filippo.gatti,didier.clouteau\}@centralesupelec.fr}


	\abstract{With the recent rise of neural operators, scientific machine learning offers new solutions to quantify uncertainties associated with high-fidelity numerical simulations. Traditional neural networks, such as Convolutional Neural Networks (CNN) or Physics-Informed Neural Networks (PINN), are restricted to the prediction of solutions in a predefined configuration. With neural operators, one can learn the general solution of Partial Differential Equations, such as the elastic wave equation, with varying parameters. There have been very few applications of neural operators in seismology. All of them were limited to two-dimensional settings, although the importance of three-dimensional (3D) effects is well known.
	
	In this work, we apply the Fourier Neural Operator (FNO) to predict ground motion time series from a 3D geological description. We used a high-fidelity simulation code, SEM3D, to build an extensive database of ground motions generated by 30,000 different geologies. With this database,  we show that the FNO can produce accurate ground motion even when the underlying geology exhibits large heterogeneities. Intensity measures at moderate and large periods are especially well reproduced. 
	
	We present the first seismological application of Fourier Neural Operators in 3D. Thanks to the generalizability of our database, we believe that our model can be used to assess the influence of geological features such as sedimentary basins on ground motion, which is paramount to evaluating site effects.}

\begin{document}

	\section{Introduction}
	Thanks to the recent increase in computational resources, high-fidelity simulations have become more affordable for diverse engineering applications. Complex physical systems can be emulated with simulations depending on many parameters. However, depending on the data available to calibrate the simulations, some parameters may be poorly constrained. Therefore, a lot of efforts are now devoted to quantifying the influence of each parameter on the global system response \parencite{hommaPhysicsbasedMonteCarlo2014, parkQuantifyingMultipleTypes2011, xiaoQuantificationModelUncertainty2019}.
	
	Physics-based simulations rely on the discretization of the underlying Partial Differential Equations (PDEs). The most common methods to solve PDEs include Finite Elements Methods, Finite Difference Methods, Finite Volume Methods, and Spectral Elements Methods. Despite this diversity, all methods suffer from high computational costs, especially for complex physical systems. Since hundreds to  thousands of simulations are necessary for uncertainty quantification analyses in Monte-Carlo Markov Chain frameworks, physics-based models cannot be used directly and faster methods need to be designed.
	
	To reduce the computational time of high-fidelity simulations, surrogate models have been developed. Classical methods, such as Gaussian processes, Kriging, Polynomial chaos, etc., depend on a limited set of parameters \parencite{abbiatiGlobalSensitivityAnalysis2021, cruz-jimenezBayesianInferenceEarthquake2018, pfortnerPhysicsinformedGaussianProcess2022, tamhidiConditionedSimulationGroundMotion2022a, sochalaModelReductionLargescale2020}. Uncertainty quantification has benefited from the development of neural networks which account for a larger number of parameters and reduce the \textit{a priori} selection of influential variables. Among standard methods, Convolutional Neural Networks (CNNs, \cite{lecunDeepLearning2015}) and Physics-Informed Neural Networks (PINNs, \cite{raissiPhysicsinformedNeuralNetworks2019}) have witnessed successful engineering applications \parencite{karniadakisPhysicsinformedMachineLearning2021, rossGeneralizedSeismicPhase2018, songVersatileFrameworkSolve2022}. However, one major drawback of these methods is the difficulty of transferring knowledge between different configurations. For example, when solving the wave equation, CNNs and PINNs are trained with a fixed velocity parameter and cannot predict anything for a different velocity value.
	
	Neural operators have been introduced to solve this issue and improve the generalizability of deep learning methods \parencite{liNeuralOperatorGraph2020, liFourierNeuralOperator2021, luLearningNonlinearOperators2021}. The paradigm of neural operators is to consider inputs and outputs as functionals. Therefore, PDEs can be solved for a large class of parameters. This work focuses on the Fourier Neural Operator (FNO) developed by \cite{liFourierNeuralOperator2021}. In seismology, the FNO has been applied for earthquake localization \cite{sunAcceleratingTimeReversalImaging2022}, to predict high-frequency terms from low-frequency simulations \cite{songHighfrequencyWavefieldExtrapolation2022}, and to solve the acoustic wave equation \cite{liSolvingSeismicWave2022, weiSmalldatadrivenFastSeismic2022}. To our best knowledge, the first application on the elastic wave equation, \textit{i.e.} solving all components of ground motion, was done by \cite{yangRapidSeismicWaveform2022, yangSeismicWavePropagation2021}. 
	
	However, all those applications were limited to one-dimensional (1D) or two-dimensional (2D) domains, while it is known that three-dimensional (3D) effects may be crucial \cite{moczoKeyStructuralParameters2018,zhuSeismicAggravationShallow2020}. Designing a 3D FNO is a technical challenge due to the memory requirements of the model. The very few studies that attempted to do so either used linear attention to reduce the dimensionality \cite{wenhuipengLinearAttentionCoupled2022} or advanced model parallelization \cite{gradyModelparallelFourierNeural2022, witteIndustryscaleCO2Flow2022}. 
	
	In this study, we conducted the first prediction of elastic waves propagation in a 3D domain. We used a FNO architecture called the U-shaped neural operator (UNO,  \cite{rahmanUshapedNeuralOperators2022}) that allows a deep architecture with ten layers. Solving a time-dependent PDE on a 3D domain would ultimately represent a 4D problem (three spatial and one temporal dimensions) which is not affordable. To circumvent this issue, we propose a dimension conversion from the input's depth dimension to the output's temporal dimension. We applied our UNO to seismological data, but our conclusions can be extrapolated to other fields involving propagation equations.

	\section{Neural Operator architecture}
	
	\subsection{Problem formulation}
	The propagation of seismic waves in a heterogeneous isotropic material can be described by the elastic wave equation
	\begin{equation}
	\rho \dfrac{\partial^2 \bm{u}}{\partial t^2} = \nabla \lambda \left( \nabla \cdot \bm{u} \right) 
	+ \nabla \mu \left[ \nabla \bm{u} + \left( \nabla \bm{u} \right)^T \right] 
	+ \left( \lambda + 2 \mu \right) \nabla \left( \nabla \cdot \bm{u} \right)
	- \mu \nabla \times \nabla \times \bm{u}
	\label{eq:elastic_equation}
	\end{equation}
	where $\rho$ is the material density, $\bm{u}$ is the displacement, and $\lambda$, $\mu$ are the Lamé parameters characterizing the material. Alternatively, the material can be described by the velocity of shear waves, $V_S$, and the velocity of compressional waves, $V_P$. In this work, we assumed that the ratio $V_P/V_S$ was constant and equal to $\alpha$ (where $\alpha$=1.7) to describe the material with the only parameter $V_S$. Therefore, equation \ref{eq:elastic_equation} can be reformulated in terms of $V_S$ as 
	\begin{equation}
	\dfrac{\partial^2 \bm{u}}{\partial t^2} = (\alpha^2-2) \nabla V_S^2 \left( \nabla \cdot \bm{u} \right) 
	+ \nabla V_S^2 \left[ \nabla \bm{u} + \left( \nabla \bm{u} \right)^T \right] 
	+ (\alpha^2 - 1) V_S^2 \nabla \left( \nabla \cdot \bm{u} \right)
	- V_S^2 \nabla \times \nabla \times \bm{u}
	\label{eq:elastic_equation_Vs}
	\end{equation}
	
	In a general framework, let us denote $D_a \subset \mathbb{R}^3$ the material domain and $\mathcal{A} = \mathcal{F}(D_a; \mathbb{R}^+)$ the space of functions describing the material property $V_S$. The displacement $\bm{u}$ intrinsically depends on four variables, $(x,y,z,t) \in D_a \times [0,T]$. However, this formulation is i) too demanding in terms of computational resources for a 4-dimensional neural network, ii) not realistic since the displacement is generally only measured with sensors at the surface. Therefore, we introduce the set $D_u = \{ (x, y, t), (x,y) \in \partial D_a^{top}, t \in [0,T] \}$ where $\partial D_a^{top}$ is the upper boundary of the material domain $D_a$ and $T>0$ is the final time. Then, we denote $\mathcal{U} = \mathcal{C}(D_u; \mathbb{R}^3)$ the space of continuous velocity functions $\bm{u}$ observed only at the surface of the domain. Note that the output space of $\bm{u}$ is $\mathbb{R}^3$ since the velocity is expressed along three components (East-West, North-South, vertical). 
	
	Then, denoting $a(\bm{x}) = V_S(\bm{x})^2$, equation \ref{eq:elastic_equation_Vs} can be expressed in the general form $\mathcal{L}(a,u)=0$ with initial conditions imposed by the seismic source and absorbing boundary conditions to represent an infinite spatial domain. In this setting, we finally aim at finding an approximation of
	\begin{equation}
	\mathcal{G} : \begin{array}{ccc}
	\mathcal{A} & \to & \mathcal{U} \\
	a & \mapsto & \bm{u}
	\end{array}
	\label{eq:operatorG}
	\end{equation}
	defined as a neural network $G_{\theta}$ depending on a set of parameters $\theta$. 
	
	\subsection{Fourier Neural Operator}
	The FNO formulation originates from the theory of kernel integral operators \parencite{huangIntroductionKernelOperator2022, liNeuralOperatorGraph2020}. It has been further modified to benefit from the computational efficiency of the Fast Fourier Transform (\cite{liFourierNeuralOperator2021}). The FNO can be decomposed into three components: 
	\begin{enumerate}
		\item an uplift sub-network $P$ that transforms the input $a$ into an abstract and higher-dimensional vector $v_0$
		\item a succession of $L$ Fourier layers transforming abstract vectors $v_{\ell} \mapsto v_{\ell+1}$, described below
		\item a projection sub-network $Q$ that maps the abstract vector $v_L$ resulting from the last Fourier layer to the physical output $\bm{u}$
	\end{enumerate}
	A Fourier layer $F_{\ell}$ can be described, with $\mathcal{F}$ denoting the Fast Fourier Transform, by 
	\begin{equation}
	v_{\ell+1} = \sigma \left( \mathcal{F}^{-1} (R_{\ell} \cdot \mathcal{F}(v_{\ell})) + W_{\ell} v_{\ell} \right)
	\label{eq:Fourier_layer}
	\end{equation}
	It first performs the convolution of the Fourier coefficients of its input vector $\mathcal{F}(v_{\ell})$ with a kernel operator $\mathcal{K}_{\ell}$ learnt during the neural network training. This is similar to the operation performed by usual convolutional layers, except that the kernel is no longer a matrix but an operator. To alleviate the computational cost, the convolution is performed in the Fourier space, where it is equivalent to multiplication. Therefore, the kernel operator $\mathcal{K}_{\ell}$ can be represented by a fixed number of Fourier coefficients $R_{\ell} := \mathcal{F}(\mathcal{K}_{\ell})$. To maintain the physical meaning throughout the process, the convolution is followed by the inverse Fourier transform $\mathcal{F}^{-1}$. Meanwhile, a bias is added through a linear operation parametrized by $W_{\ell}$. Finally, a non-linear activation function $\sigma$ is applied. 
	
	One major advantage of neural operators is their discretization-invariance property. Indeed, the weights $R_{\ell}$ depend on the number of Fourier coefficients one chooses to select in each Fourier layer and not on the dimension of $v_{\ell}$. Therefore, the neural operator can be trained with moderate dimensions and, once trained, be applied to higher-resolution data to highlight small-scale features.

	\subsection{Model used in this work}
	To increase the complexity of the neural network, and hence its expressivity, while avoiding the common problem of vanishing/exploding gradients, the UNO has been proposed by \cite{rahmanUshapedNeuralOperators2022}. It arranges the Fourier layers in an encoder-decoder structure to allow skip connections from one Fourier layer in the encoder to its symmetric layer in the decoder. Our model comprises 8 Fourier layers, 4 in the encoder and 4 in the decoder. 
	
	\begin{figure}[h]
		\centering
		\includegraphics[width=\textwidth]{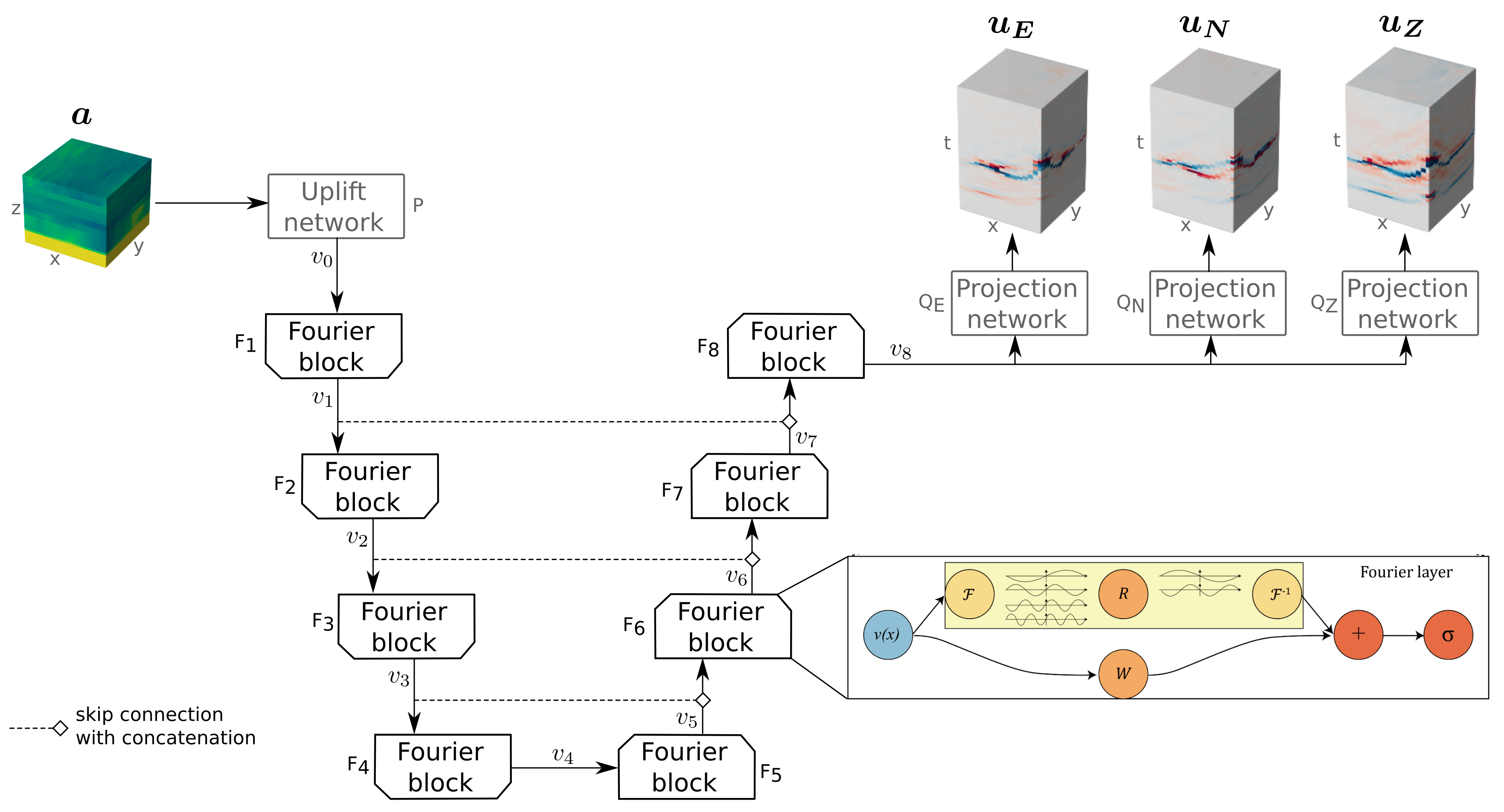}
		\caption{Architecture of our UNO. The input (3D geology $a$) is uplifted by the sub-network $P$, then transformed trough 8 Fourier layers $F_1, \cdots, F_8$. Finally, three sub-networks $Q_E$, $Q_N$, and $Q_Z$ project the components of the velocity $u_E$, $u_N$, $u_Z$. The detail of the sixth Fourier block is shown in the right corner (image reproduced from \cite{liFourierNeuralOperator2021}). Dotted lines show the skip connections with a concatenation of inputs.}
		\label{fig:UNO}
	\end{figure}
	
	In addition to adding skip connections to the original FNO, the UNO modifies the dimensions of vectors $v_{\ell}$ in each Fourier layer. In the encoder part, the physical dimensions are reduced from $64~\times~64~\times~64$ to $8~\times~8~\times~8$ while they are increased to $64~\times~64~\times~128$ in the decoder part. 
	
	The uplift sub-network $P$ is made of two fully connected layers. It takes as input the geological description $a$ concatenated with a positional encoding of coordinates in $D_a$. With this concatenation, $P$ creates an abstract vector $v_0$ with the same dimensions as $a$ and 16 additional channels. At the end of the model, there are three projection operators $Q_E$, $Q_N$, and $Q_Z$, each made of two fully connected layers. Each operator projects the vector $v_L$ onto the respective velocity components $u_E$, $u_N$, $u_Z$.

	\section{Data}
	The UNO is trained in a fully supervised manner. Therefore, the training requires pairs $(a, \bm{u})$. Geological inputs were built from random fields to form a large and general database (Section \ref{sec:data_geology}). Then, finite elements simulations were performed to obtain the velocity fields at the surface (Section \ref{sec:data_SEM}).

	\subsection{Geological database}
	\label{sec:data_geology}
	Our geological database represents 3D S-wave velocity fields in a $9.6~\text{km}~\times~9.6~\text{km}~\times~9.6~\text{km}$ cube. This database is built from layered random fields to provide a general decomposition basis to any 3D material \cite{lehmannMachineLearningOpportunities2022}. Each geology contains a homogeneous bottom layer with constant velocity $V_S$=4500 m/s. On top of this layer, there are 1 to 6 layers whose thickness and mean value were randomly chosen. The mean value follows a uniform distribution $\mathcal{U}$(1785 m/s; 3214 m/s). There is no additional constraint, meaning that velocity inversion may be present. 
	
	Then, heterogeneities were added independently inside each layer with von Karman random fields which are known to best represent crustal heterogeneities. The correlation length was randomly chosen between 1.5km and 6km, and the coefficient of variation followed a normal distribution $\mathcal{N}(0.2, 0.1)$ of mean 0.2 and standard deviation 0.1. Finally, the velocities were clipped to the interval [1071 m/s; 4500 m/s] to ensure realistic values and an appropriate discretization of wave lengths. 
	
	For the neural operator training, geologies were represented as $64 \times 64 \times 64$ matrices. The geological database amounts to 30Go.

	\subsection{Finite elements 3D simulations}
	\label{sec:data_SEM}
	To obtain ground motion time series, we used SEM3D, a High-Performance Computing (HPC) code based on the Spectral Element Method \parencite{komatitschIntroductionSpectralElement1999, touhamiSEM3D3DHighFidelity2022}. Each geology was discretized to a hexahedral mesh with elements of size 300m. A seismic source was placed in the middle of the bottom layer at position (4.8km, 4.8km, -8.4km). It is parametrized as a moment tensor with strike~=~50\textdegree, dip~=~45\textdegree, and rake~=~88\textdegree. Its source time function is given by $t \mapsto 1 - \left(1 + \frac{t}{\tau} \right) e^{-\frac{t}{\tau}}$ with $\tau$ = 0.127 s. These parameters correspond to the Le Teil earthquake (France, 2019, \cite{delouisConstrainingPointSource2021}).
	
	Each simulation runs for 20s and can accurately reproduce frequencies up to 5 Hz. The generation of 30,000 simulations represents 1.6 million CPU hours (50 min per simulation on 64 CPUs). During the simulation, ground motion velocities were recorded on a grid of 256 virtual sensors equally spaced at the surface (the space between two consecutive sensors was 600m). The acquisition frequency is 20 Hz. 
	
	For the neural operator training, we used the ground motion time series between 1s and 7.4s (128 time points). In addition, we performed a 2D spatial interpolation to refine ground motion to a $64 \times 64$ spatial grid. The velocity database amounts to 176 Go.

	\subsection{Neural operator training}
	The database of 30,000 instances was split into 90\% training data and 10\% validation data. The geological inputs were normalized to a Gaussian with mean 0 and standard deviation 0.25. No normalization was applied to the outputs.
	
	The neural operator has been trained with the Adam optimizer for 110 epochs. The learning rate was initially $10^{-3}$ and was reduced by 0.5 when the validation loss did not improve for 20 epochs. The activation function was the relu function and the loss function was defined as the Mean Absolute Error (MAE). In total, the neural operator contains 87 million parameters. The training took 11 hours on 4 Nvidia A100 GPUs.

	\section{Results}
	\subsection{General prediction results}
\begin{figure}[h]
		\centering
		\includegraphics[width=0.7\textwidth]{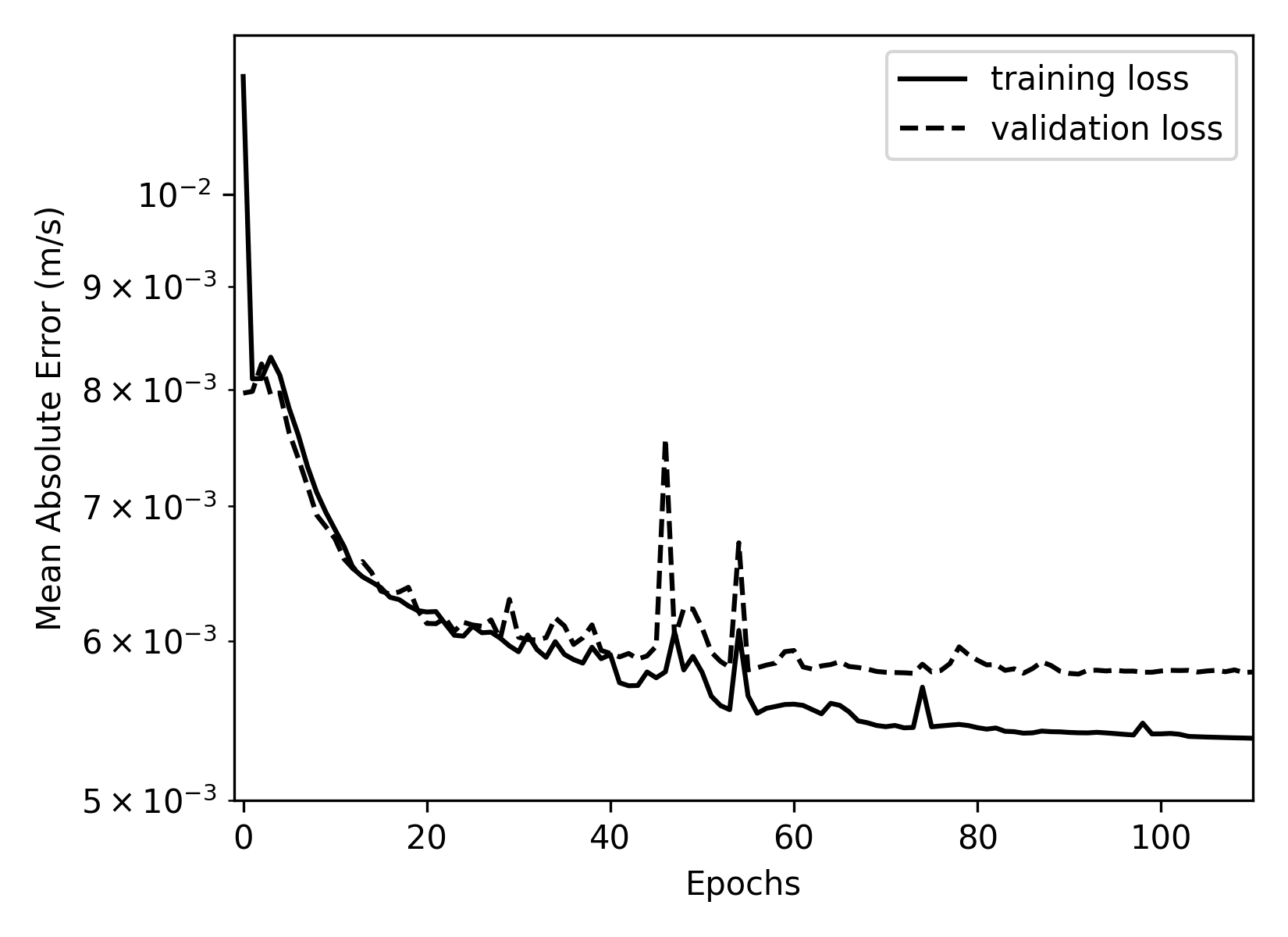}
		\caption{Evolution of the training loss (line) and the validation loss (dashed line) as a function of epochs. The loss is summed over the three components.}
		\label{fig:loss_evolution}
	\end{figure}

\begin{figure}[h]
		\centering
		\includegraphics[width=\textwidth]{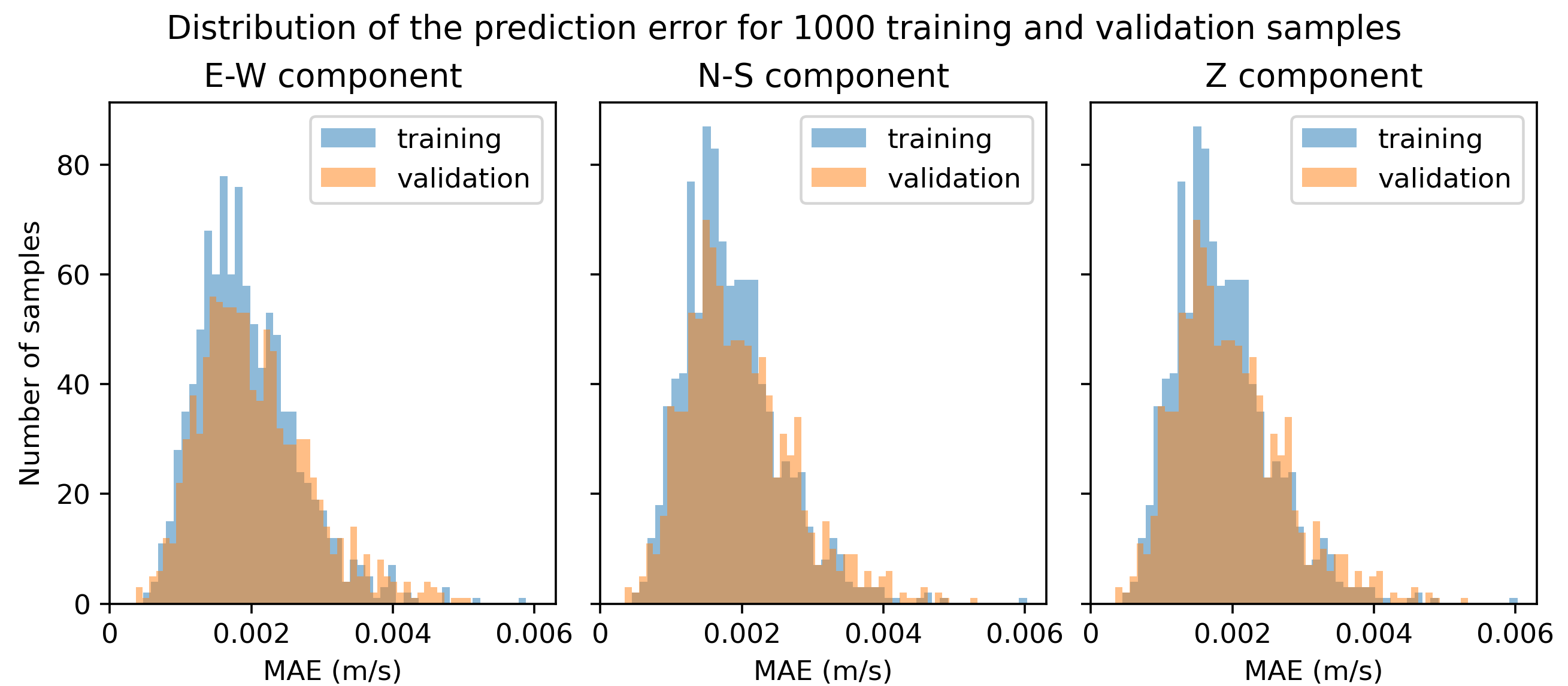}
		\caption{For 1000 elements of the training dataset (blue) and validation dataset (orange), distribution of the Mean Absolute Error (MAE) between the neural operator prediction and the ground truth. Each subpanel shows one velocity component (E-W: East-West), (N-S: North-South), (Z: vertical).}
		\label{fig:loss_components}
	\end{figure}

	Figure \ref{fig:loss_evolution} shows the convergence of the training and validation loss functions. The exponential decreasing curve illustrates the quality of the neural operator learning. At the end of the training, the validation loss is slightly higher than the training loss, which may indicate a slight overfitting of the data. However, when comparing the distribution of the prediction error for 1,000 instances in the training and validation datasets, it appears that the distributions are very similar (Figure \ref{fig:loss_components}). Therefore, if overfitting is present, it should have a minimal effect on the validation results' accuracy.
	
	Additionally, Figure \ref{fig:loss_components} shows that the prediction accuracy is similar for the three components (East-West, North-South, vertical). This means that the projection sub-networks $Q_E$, $Q_N$, and $Q_Z$ are sufficient to capture the specificities of each velocity component from the last abstract vector $v_8$. Therefore, the following results are presented only for the East-West component, and their interpretation can be easily transferred to the other components.
	
	\subsection{Ground motion prediction for validation data samples}
\begin{figure}[p]
		\centering
		\begin{subfigure}[b]{0.65\textwidth}
			\centering
			\includegraphics[width=0.7\textwidth,trim=250 100 0 100, clip=true]{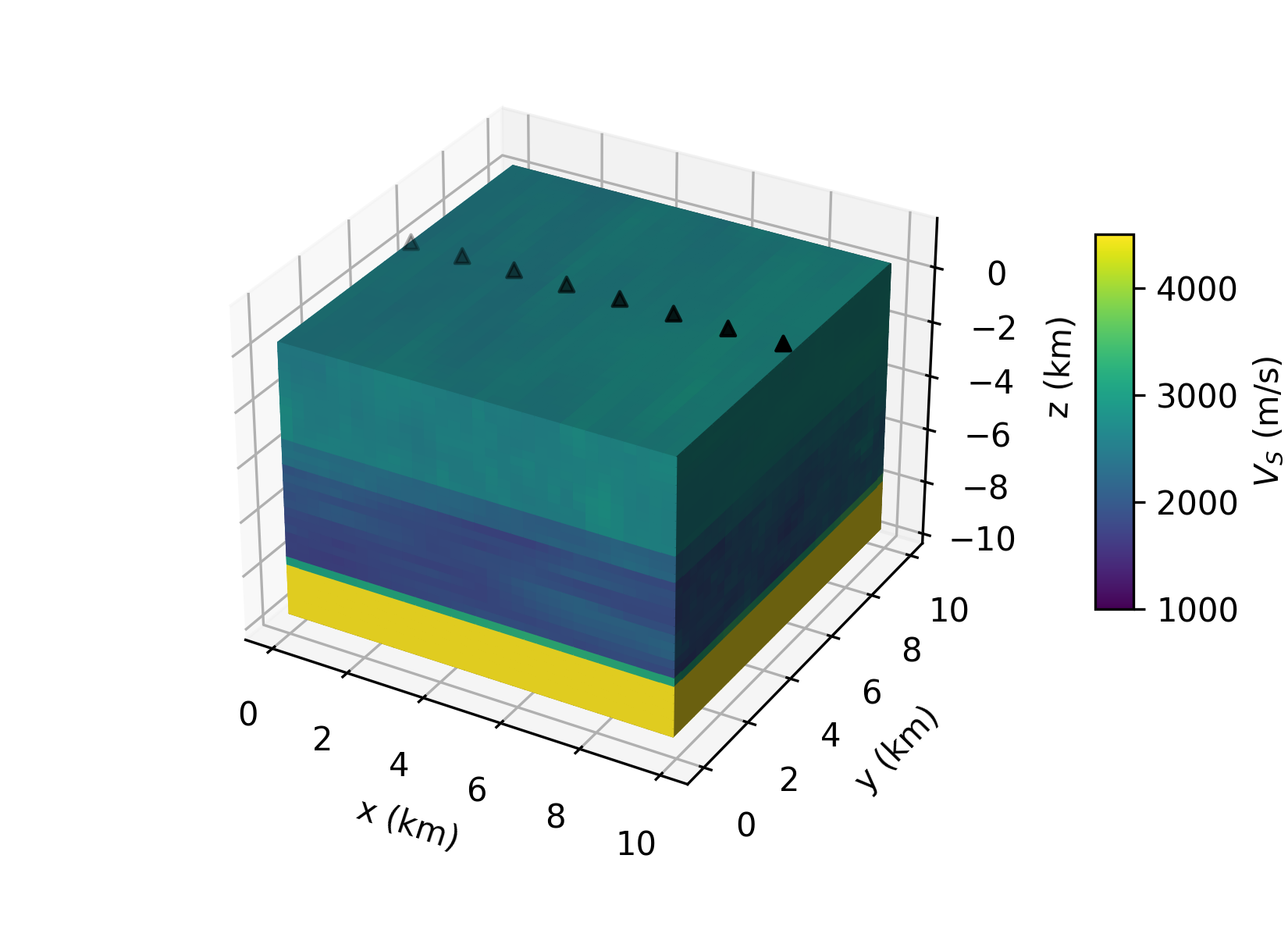}
			\caption{3D geological description (S-wave velocity $V_S$). The black triangles represent the sensors where the traces are shown in Figure \ref{fig:val87_traces}.}
			\label{fig:val87_Vs}
		\end{subfigure}
		\hfill
		\begin{subfigure}[b]{0.56\textwidth}
			\includegraphics[width=\textwidth]{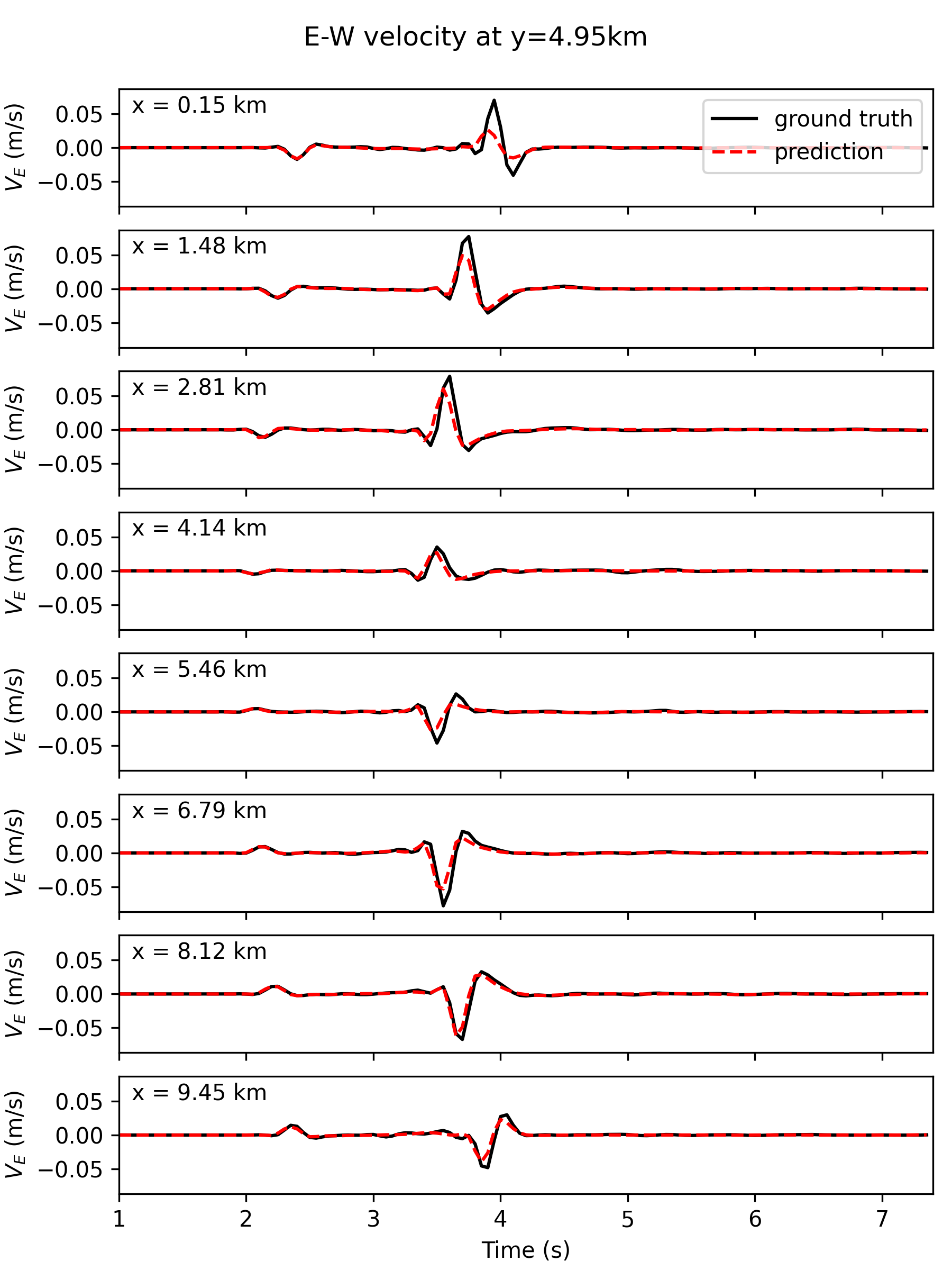}
			\caption{E-W ground velocity time series at 8 sensors aligned at y=4.95km and represented in Figure \ref{fig:val87_Vs}.}
			\label{fig:val87_traces}
		\end{subfigure}
		\hfill
		\begin{subfigure}[b]{0.41\textwidth}
			\includegraphics[width=\textwidth]{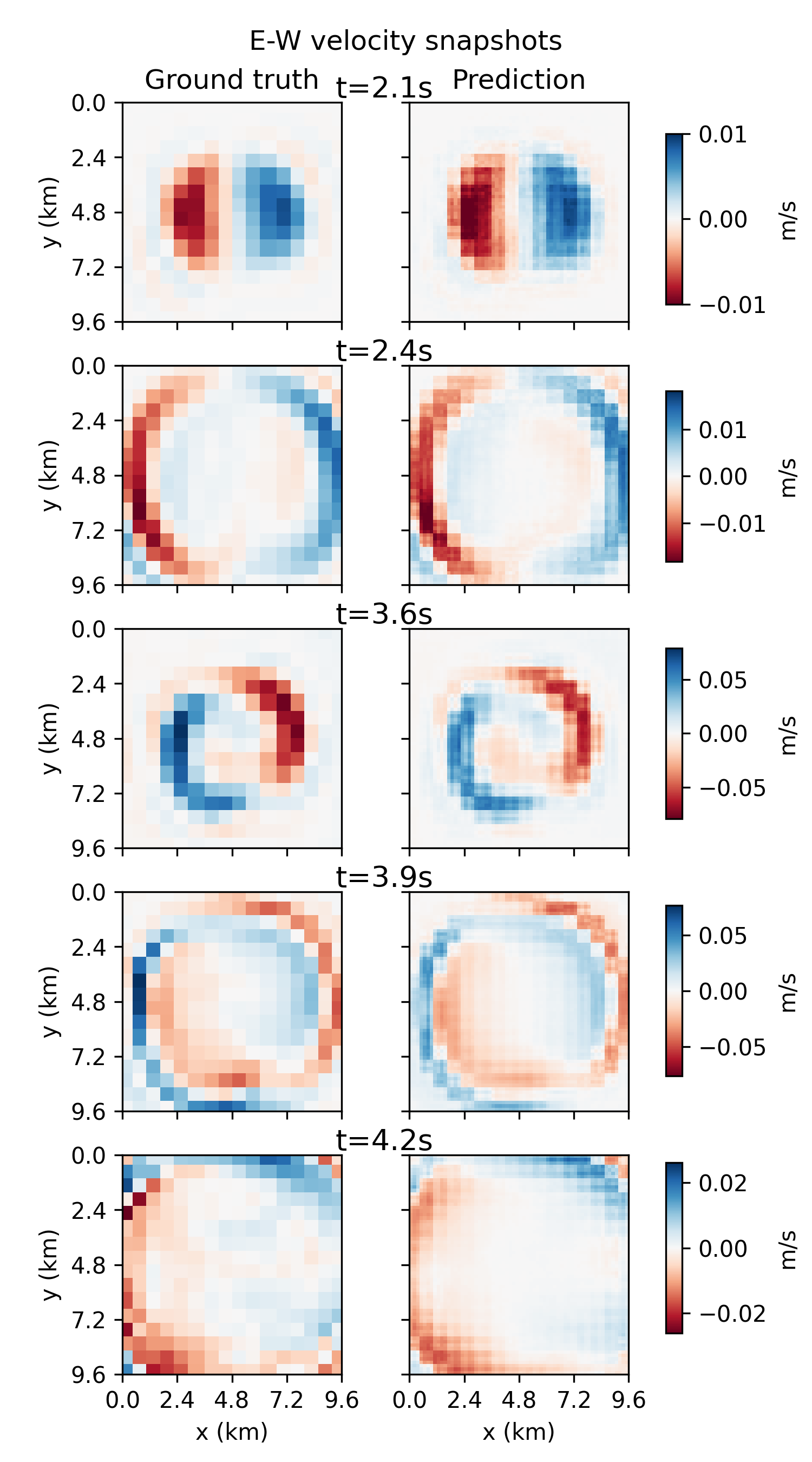}
			\caption{Comparison of reference velocities (left column) and velocities predicted by the neural operator (right column) for 5 time points.}
			\label{fig:val87_snapshots}
		\end{subfigure}
		\caption{Comparison of simulations (considered as ground truth) and neural operator predictions for one geology in the validation dataset.}
		\label{fig:val87}
	\end{figure}

	Figure \ref{fig:val87} shows the neural operator prediction for one geology in the validation dataset. This geology has four layers with large velocity contrasts between layers (Figure \ref{fig:val87_Vs}). The MAE on the E-W velocity component is $7.8 \: 10^{-4}$m/s. The time evolution at individual snapshots shows that the neural operator can accurately capture the time arrival of P-waves (the first arrival) and S-waves (the second and most prominent peak), despite their different amplitudes (Figure \ref{fig:val87_traces}). In addition, the velocity magnitude is also spatially close to the reference, as seen on the snapshots in Figure \ref{fig:val87_snapshots}. 
	
	We found that the neural operator tends to underestimate the Peak Ground Velocity (PGV). This is visible in Figure \ref{fig:val87_traces} where the prediction peaks are generally smaller than the ground truth. 
	
\begin{figure}[p]
		\centering
		\begin{subfigure}[b]{0.65\textwidth}
			\centering
			\includegraphics[width=0.7\textwidth,trim=250 100 0 100, clip=true]{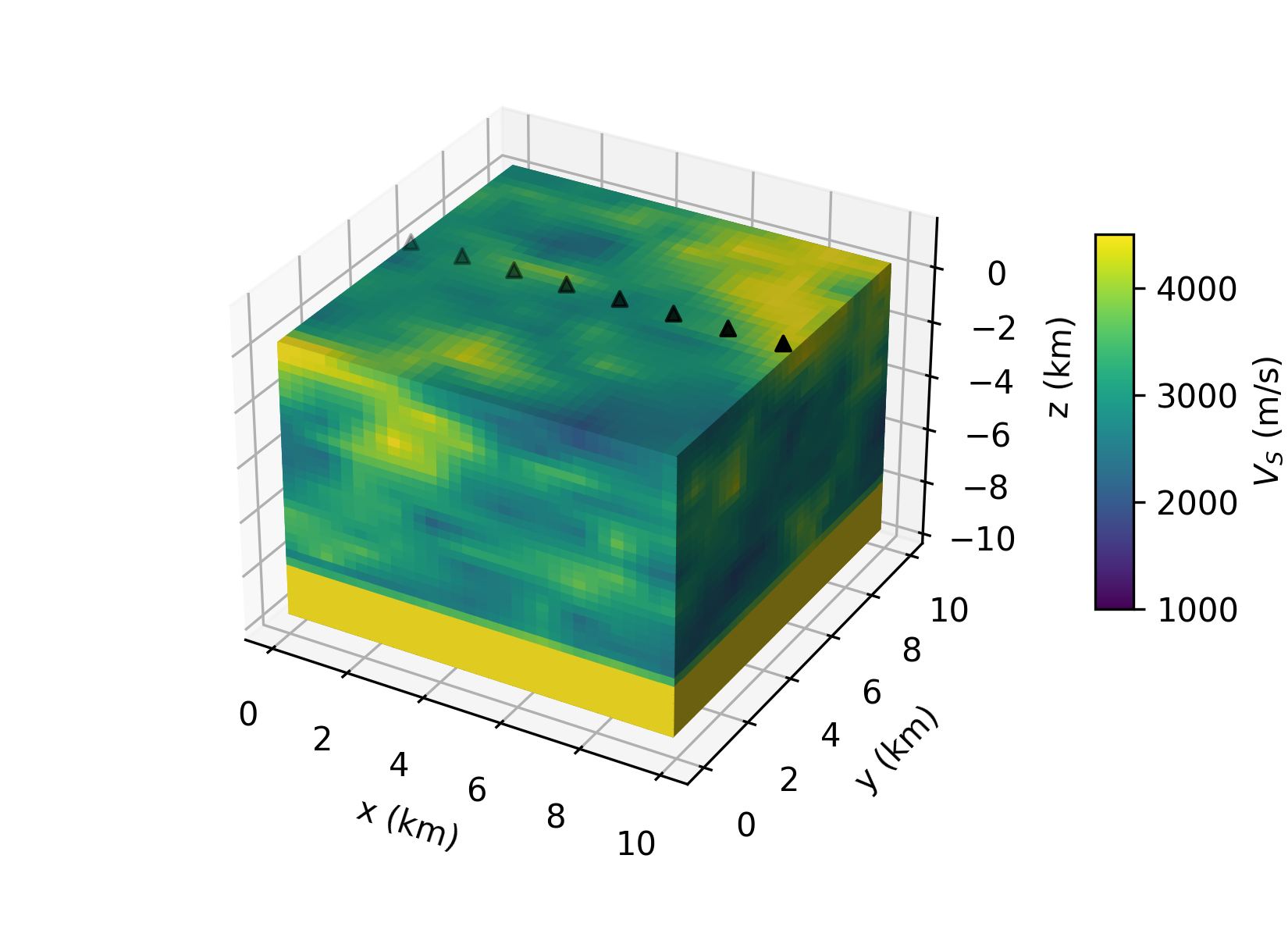}
			\caption{3D geological description (S-wave velocity). The black triangles represent the sensors where the traces are shown in Figure \ref{fig:val7_traces}.}
			\label{fig:val7_Vs}
		\end{subfigure}
		\hfill
		\begin{subfigure}[b]{0.56\textwidth}
			\includegraphics[width=\textwidth]{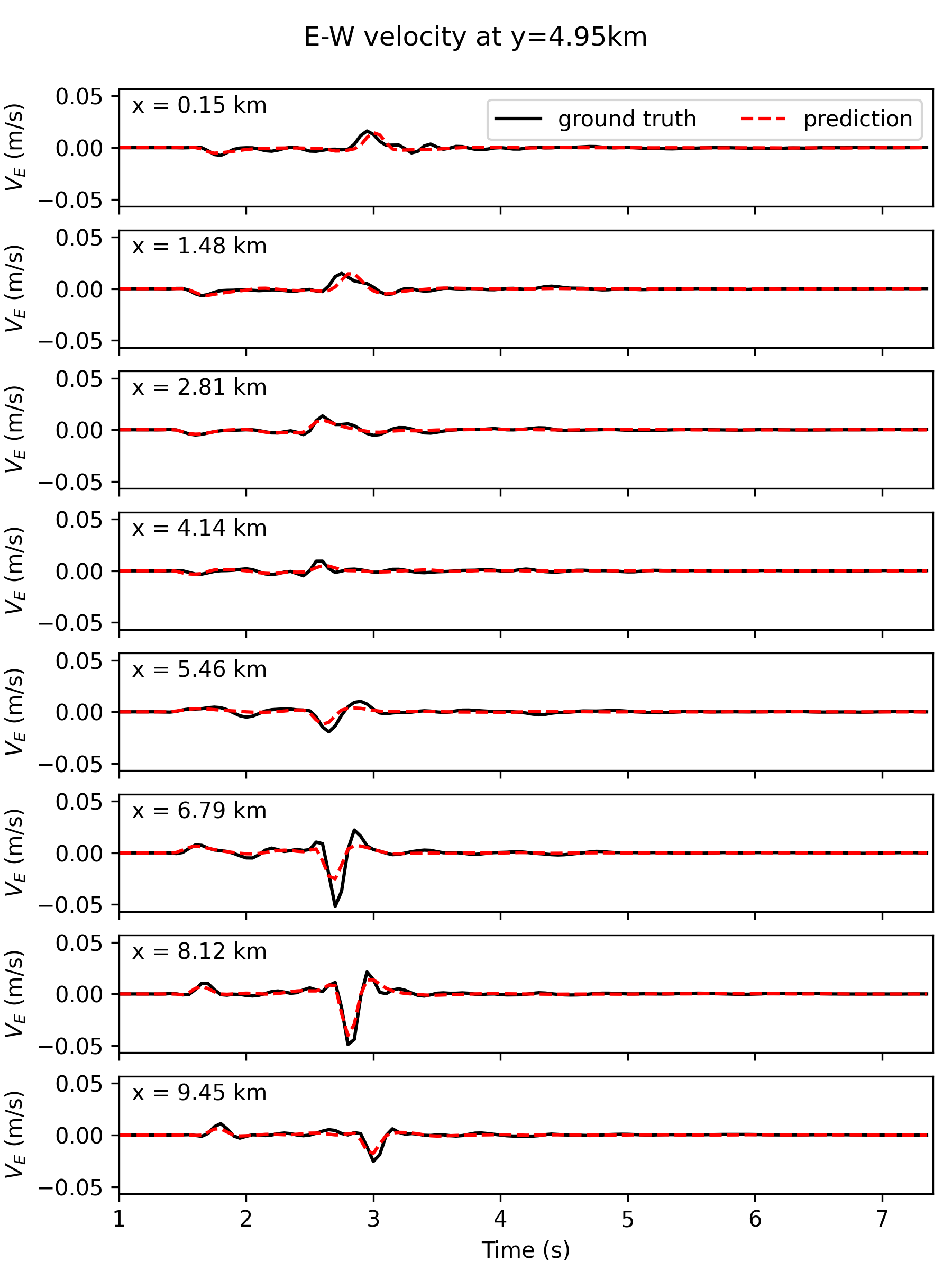}
			\caption{E-W ground velocity time series at 8 sensors aligned at y=4.95km and represented in Figure \ref{fig:val7_Vs}.}
			\label{fig:val7_traces}
		\end{subfigure}
		\hfill
		\begin{subfigure}[b]{0.41\textwidth}
			\includegraphics[width=\textwidth]{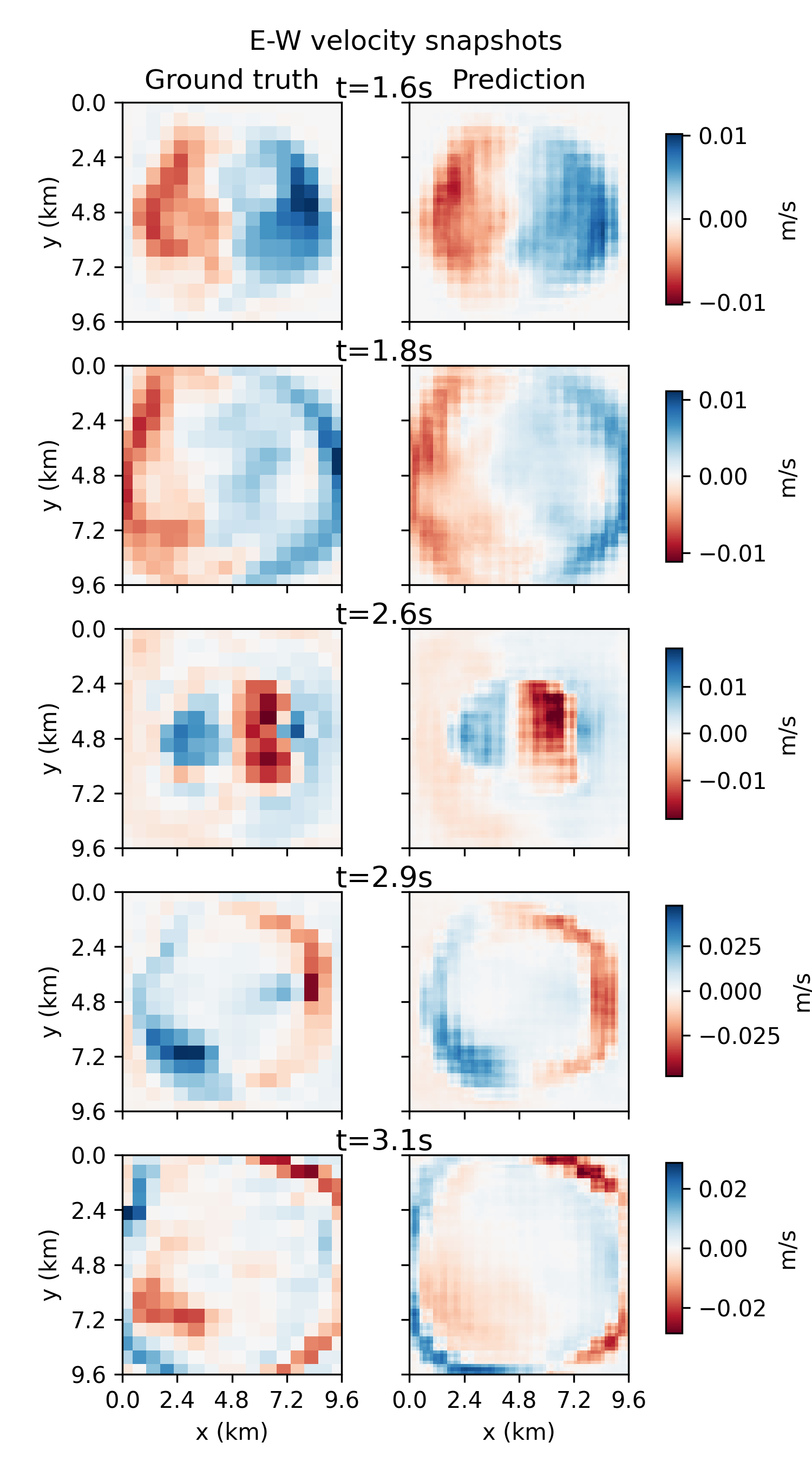}
			\caption{Comparison of reference velocities (left column) and velocities predicted by the neural operator (right column) for 5 time points.}
			\label{fig:val7_snapshots}
		\end{subfigure}
		\caption{Comparison of simulations (considered as ground truth) and neural operator predictions for one geology in the validation dataset.}
		\label{fig:val7}
	\end{figure}

	Figure \ref{fig:val7} shows that predictions are also accurate when the neural operator input is a heterogeneous geology like depicted in Figure \ref{fig:val7_Vs}. Heterogeneities diffracted the seismic waves, leading to more variations in the velocity time series. However, the wave arrival times are still accurately captured by the neural operator (Figure \ref{fig:val7_traces}), and the spatial distribution of ground motion is very satisfactory (Figure \ref{fig:val7_snapshots}).

\begin{figure}[h]
		\centering
		\includegraphics[width=0.8\textwidth]{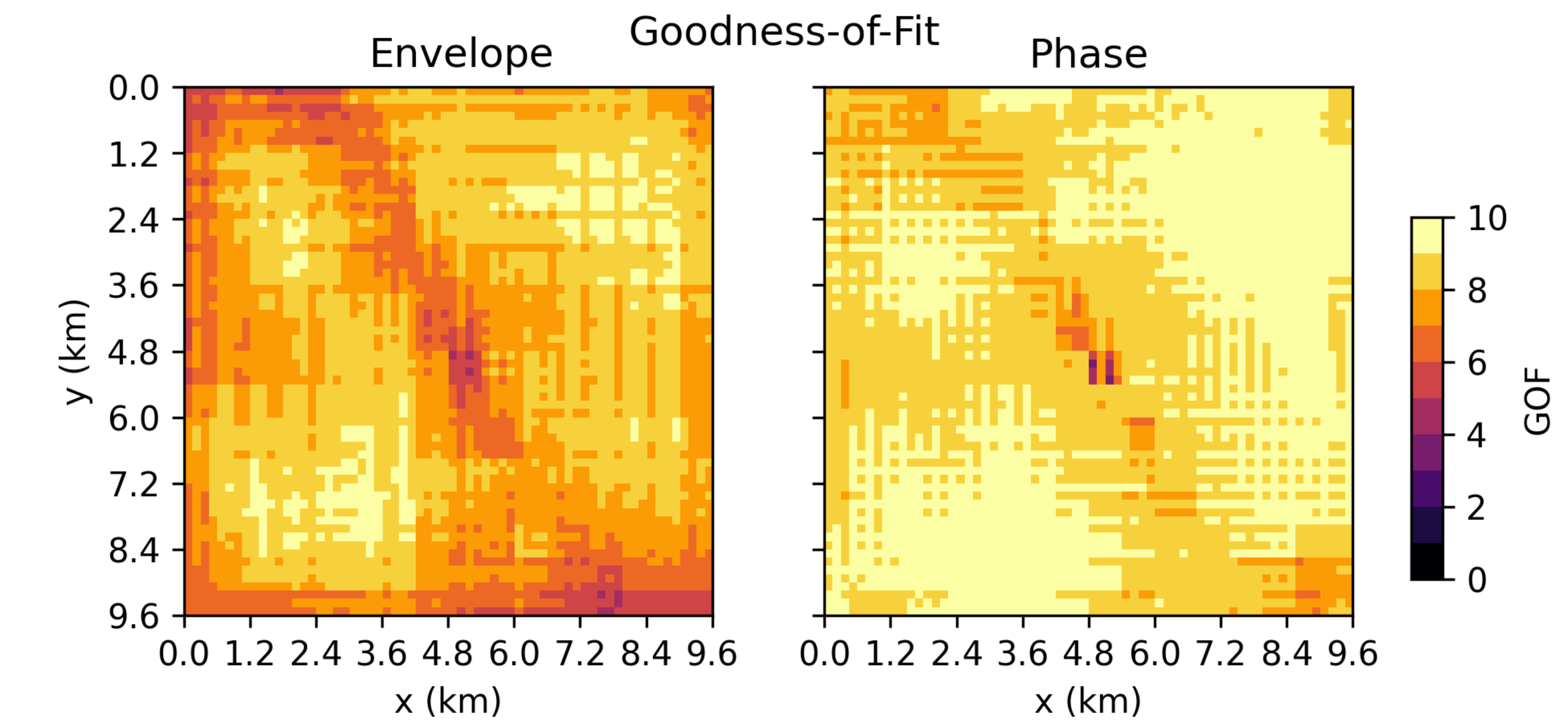}
		\caption{Goodness-of-Fit (GOF) criteria for the envelope and phase of all time series predicted with the heterogeneous geology (Figure \ref{fig:val7_Vs}). 10 means a perfect agreement.}
		\label{fig:val7_gof}
	\end{figure}

	Quantitatively, one can compute the Goodness-of-Fit criteria \parencite{kristekovaTimefrequencyMisfitGoodnessoffit2009}. This scale evaluates the difference between two time series regarding envelope and phase. Scores above 6 are generally considered good and above 8 considered very good (10 is a perfect match). Figure \ref{fig:val7_gof} shows the GOF corresponding to the predictions with the heterogeneous geology in Figure \ref{fig:val7}. One can observe that the envelope GOF is larger than 6 for 95\% of the points, and the phase GOF is larger than 8 for 92\% of points. These scores assess the quality of the UNO prediction. In addition, the better agreement on the phase confirms that the predictions are more accurate for the wave arrival times than for the peaks' magnitude.
	
\begin{figure}[h]
		\centering
		\includegraphics[width=0.9\textwidth]{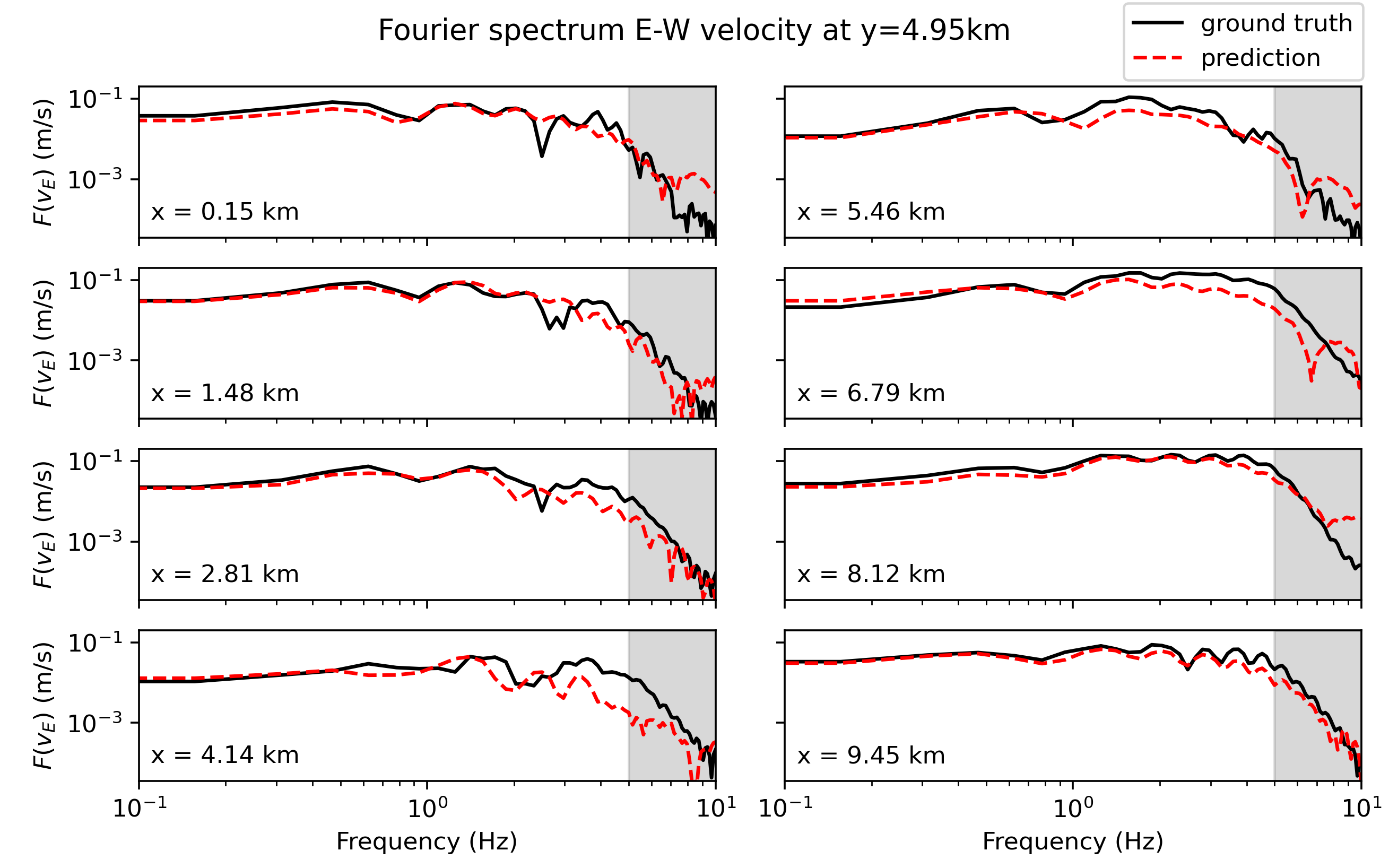}
		\caption{Amplitude of Fourier coefficients for reference velocity time series (black line) and time series predicted by the neural operator (dashed red line). The grey area represents the maximum frequency (5 Hz) where the numerical simulations are valid.}
		\label{fig:val7_fourier}
	\end{figure}

	When examining the velocity fluctuations in Figure \ref{fig:val7_traces}, it appears that the smallest ones are generally missing from the neural operator prediction. This can also be observed in Figure \ref{fig:val7_fourier} where the Fourier coefficients are shown as a function of the frequency. Although the prediction is very close to the ground truth at low frequency, the disagreement increases above 1 Hz. However, the amplitude decay has approximately the same slope for the prediction and the reference signal, meaning that the neural operator lacks accuracy on high-frequency variations but does not introduce data filtering.  
	
	\section{Discussion and conclusion}
	We used a database of 3D heterogeneous geologies based on random fields to simulate the propagation of seismic waves. We showed that the UNO is able to learn the relationship between the geology and the surface ground motion represented by 3-component velocity time series. Therefore, our model maps the depth dimension of the geology to the time dimension of the velocity. This is an efficient way to model the 3D elastic wave equation while limiting the memory requirements of the model. 
	
	Contrary to a common belief in machine learning that it is better to start with easy tasks, we found that the training significantly improved when predicting the three components instead of a single one. This may be explained by the fact that adding two projection sub-networks $Q_N$ and $Q_Z$ does not significantly increase the number of parameters in the model but triples the amount of training data. Therefore, predicting multiple components was beneficial for the deep learning model. 
	
	The generalization gap between the validation loss and the training loss (Figure \ref{fig:loss_evolution}) could probably be reduced by using a larger database. Indeed, we observed that the validation loss was higher when using 20,000 samples instead of 30,000. Due to the computational cost of generating the database with high-fidelity simulations, we accepted this trade-off as a first step to validate the 3D UNO. 
	
	Despite this limitation, the predictions on validation data (Figures \ref{fig:val87} and \ref{fig:val7}) show the accuracy of our model. It cannot be used at this point to predict maximum values since they tend to be underestimated. This reflects the findings of \cite{sunAcceleratingTimeReversalImaging2022} that amplitudes are more difficult to predict than phases. However, the amplitude accuracy should be improved by using more training data. The network architecture can also be refined, especially by increasing the number of Fourier modes in each block. This should allow a better representation of high-frequency variations. Nevertheless, the wave arrival times are already accurate despite geological heterogeneities that created dispersion. This is important for future applications in earthquake early warning, where time predictions are crucial. 	
	
	\clearpage 
	
	\printbibliography

\end{document}